\documentclass{article}

\usepackage{arxiv}

\usepackage[utf8]{inputenc} % allow utf-8 input
\usepackage[T1]{fontenc}    % use 8-bit T1 fonts
\usepackage{hyperref}       % hyperlinks
\usepackage{url}            % simple URL typesetting
\usepackage{booktabs}       % professional-quality tables
\usepackage{amsfonts}       % blackboard math symbols
\usepackage{nicefrac}       % compact symbols for 1/2, etc.
\usepackage{microtype}      % microtypography
\usepackage{lipsum}
\usepackage{graphicx}
\usepackage{multirow}
\usepackage{diagbox}
\usepackage{subfigure}
\usepackage{amsmath}
\usepackage{color}
\usepackage{float}
\usepackage{verbatim}
\usepackage[justification=centering]{caption}

\title{An Improving Framework of regularization for Network Compression}

\author{
  E Zhenqian\thanks{email: 18110980004@fudan.edu.cn} \\
  School of Data Science\\
  Fudan University\\
  Shanghai 200000, China\\
   \And
 Gao Weiguo\thanks{email: wggao@fudan.edu.cn}\\
  School of Data Science \\
  Department of Mathematical Science\\
  Fudan University\\
}

\begin{document}
\maketitle
\tableofcontents

\newpage
\begin{abstract}
Deep Neural Networks have achieved remarkable success depending on the developing high computation capability of GPUs and large-scale datasets with increasing network depth and width. However, due to the expensive computation and intensive memory, researchers have concentrated on designing compression methods in recent years. In this paper, we propose an improving framework of partial regularization rather than the original form of penalizing all parameters. It is reasonable and feasible with the help of permutation invariant property of neural network. Experimental results show that since the objective functions contain less number of entries in regularization terms, the computational costs in both training and test phases are reduced. As we expect, the running time decreases in all experiments we have tested. Surprisingly, partial regularization framework brings improvements such as higher classification accuracies in both training and testing stages on multiple benchmark datasets. What's more, we analysed the results and draw a conclusion that the optimal network structure must exist and depend on the input data.
\end{abstract}


\section{Introduction}
Deep neural networks have attracted lots of attention for their great success in many tasks since they contain millions or even billions of parameters, especially in the fully-connected layers. It has been shown expensive because of the required enormous memory and time. As is known to all, most of the parameters in such over-parametric networks are redundant and not equally important. Therefore, compact architectures could do as good a job as the deep ones. Moreover, the complexity of neural networks increases such that the training error reduces but the gap between training and test error is too large. This makes it more prone to overfitting. On the other hand, shallow networks may struggle to handle non-linearities as effectively as deeper ones because it is unlikely to fit complicated high dimensional functions. Hence, model compression for deep architectures, or more precisely determining the best number of parameters, is indeed a challenging but necessary task that should be studied. In recent years, it has already made much progress by the deep learning community. Those works do serve to reduce the storage and computational costs.\\
When considering the model compression task, one way is to impose the group sparsity regularization to automatically determine the number of neurons and select the remaining ones in each layer of a deep neural network simultaneously as the machine is learning. In fact, once a neuron is removed from the network, the outgoing weight vector grouping defined in the paper \cite{ochiai2017automatic} will have no effect on the final output. In this case, we can remove the entire group as each group is defined to act on all the outgoing weights from the belonging neuron. Setting these parameters to zero amounts to cancelling the influence of a particular neuron and thus removing it entirely. As a consequence, group regularization method and its variant don't require training an initial and large neural network as a pre-processing step. They do not depend on the success of learning a redundant network to later reduce its parameters, but instead jointly learn the number of relevant neurons in each layer and the parameters of these neurons.\\
Roughly speaking, the wider and deeper the network, the higher its performance is. So obviously, there are two ways to increase the complexity/capacity of neural networks, one is adding hidden layers and the other is adding hidden units. However, in order to solve the problem efficiently and not to be trapped in such overfitting dilemma, scientists always need to do a trade-off between the depth and width when designing the network structure. A recent deduction from researchers is that increasing the depth is more efficient since it not only increases the number of units but also the embedding depths of the functions. Therefore, the over-parametric phenomenon is largely due to the width, at least in our experiments. If so, we don't need to penalty all parameters in each layer since any whole layer is not required to be removed. Our main contribution in this paper is to introduce a general framework of partial regularization to improve the performances of the existing regularization methods. Experimental results demonstrate the effectiveness of our partial regularization framework operating on fully connected layers only. We compare the performances of the modified methods with some previous works on multiple famous datasets and find that our improving framework does have some advantages. It results in an improvement on some metrics that we are interested in on both training and test stages. At the same time, it reduces the computation costs since less entries are involved in the calculation of the regularization terms. In short, applying partial regularization on network parameters is more powerful than full regularization and permutation invariant property in neural networks provides us efficiency in realizing our idea.

\section{Model Compression Methods}
In this section, we briefly introduce several kinds of famous algorithms in model compression, along with their advantages and disadvantages following the survey paper \cite{cheng2017survey} respectively.

\subsection{Parameter Pruning and Sharing}
Parameter pruning is to remove the redundant parameters with less information in the trained deep neural network. Those parameters are not crucial to the performance. Therefore, pruning helps to reduce the number of parameters and then speed up the computation with less storage. Furthermore, it is effective in addressing the overfitting problem.\\
In the late 20th century, the Optimal Brain Damage \cite{lecun1990optimal} and the Optimal Brain Surgeon \cite{hassibi1993second} methods were proposed to remove connections with low importance based on the Hessian of the loss function. In recent years, researchers concentrate on how to prune redundant, non-informative weights in a pre-trained CNN model. For example, Srinivas and Babu \cite{srinivas2015data}  proposed a data-free pruning method to remove redundant neurons using back propogation. Han et al. \cite{han2015learning} proposed a method to reduce the storage and computation required by neural networks without affecting their accuracy by learning only the important connections.
Parameter sharing is to create a mapping that helps to share some parameters within the same data. Quantization which replace full-precision weights by low bit ones is the most popular method in this area. Chen et al. \cite{chen2015compressing} proposed a HashedNets model that used a low-cost hash function to group weights into hash buckets for parameter sharing. The deep compression method in \cite{HanMao16} removed the redundant connections and quantized the weights, and then used Huffman coding to encode the quantized weights. In \cite{ullrich2017soft}, a simple regularization method based on soft weight-sharing was proposed, which included both quantization and pruning in one simple (re-)training procedure.\\
There is also growing interest in training compact CNNs with sparsity constraints. Those sparsity constraints are typically introduced in the optimization problem as $l_0$ or $l_1$-norm regularizers. The work in \cite{lebedev2016fast} imposed group sparsity constraint on the convolutional filters to achieve structured brain Damage, i.e., pruning entries of the convolution kernels in a group-wise fashion. In \cite{zhou2016less}, a group-sparse regularizer on neurons was introduced during the training stage to learn compact CNNs with reduced filters. Wen et al. \cite{wen2016learning} added a structured sparsity regularizer on each layer to reduce trivial filters, channels or even layers. In the filter-level pruning, all the above works used $l_{2,1}$-norm regularizers. The work in \cite{li2016pruning} used $l_1$-norm to select and prune unimportant filters.\\
There are some potential issues of the pruning and sharing works. First, pruning with $l_1$ or $l_2$ regularization requires more iterations to converge. What's more, all pruning criteria require manual setup of sensitivity for layers, which demands fine-tuning of the parameters and could be cumbersome for some applications.

\subsection{Low-Rank Factorization}
The key idea of low-rank factorization is to adopt matrix or tensor decomposition in deep neural networks. The low-rank approximation was done layer by layer. The parameters of one layer were fixed after it was done, and the layers above were fine-tuned based on a reconstruction error criterion.\\
For the convolution kernels, it can be viewed as a 4D tensor. There is a significant amount of redundancy in 4D tensor, which contributes the bulk of all computations in deep CNNs. In \cite{lebedev2014speeding}, they proposed canonical Polyadic (CP) decomposition which used nonlinear least squares and replace the original convolutional layer with a sequence of four convolutional layers with small kernels. In \cite{tai2015convolutional}, they proposed a new algorithm for computing the low-rank tensor decomposition and a new method for training low-rank constrained CNNs from scratch.\\
Considering the fully-connected layer, it can be view as a 2D matrix. There are several classical works on exploiting low-rankness in fully connected layers. For instance, Misha et al. \cite{denil2013predicting} reduced the number of dynamic parameters in deep models using the low-rank method. \cite{sainath2013low} explored a low-rank matrix factorization of the final weight layer in a DNN for acoustic modeling. Besides, dropout in \cite{srivastava2014dropout} is to set a randomly selected subset of activations to zero within each layer from the neural network during training.\\
However, the implement of low-rank factorization is not that easy since it contains decomposition operation, which is computationally expensive. Another issue is that current methods perform low-rank approximation layer by layer, and thus cannot do global model compression, which is important because different layers hold different information. Finally, factorization requires extensive model retraining to achieve convergence when compared to the original model.

\subsection{Compact Convolutional Filters}
CNNs are parameter efficient due to exploring the translation invariant property of the representations to input image, which is the key to the success of training very deep models without severe over-fitting. The key idea is to replace the loose and over-parametric filters with compact blocks and therefore helps to reduce the computation cost. It has been proved to significantly accelerate CNNs on several benchmarks. For example, decomposing $3 \times 3$ convolution into two $1 \times 1$ convolutions was used in \cite{szegedy2017inception}, which achieved state-of-the-art acceleration performance on object recognition. SqueezeNet \cite{wu2017squeezedet} was proposed to replace $3 \times 3$ convolution with $1 \times 1$ convolution, which created a compact neural network with about 50 fewer parameters and comparable accuracy when compared to AlexNet.\\
There are some drawbacks of these methods. Firstly, they cannot achieve competitive performance for narrow or special ones (like GoogleNet, Residual Net). Secondly, it is difficult to be used together with other technologies for improving the compression performance. Last but not least, it cannot generalize so well and may cause unstable results for some datasets.

\subsection{Knowledge Distillation}
The main idea of Knowledge Distillation based approaches is to shift knowledge from a large teacher model into a small one by learning the class distributions output via softened softmax. In \cite{hinton2015distilling}, they introduced a KD compression framework which compressed an ensemble of deep networks (teacher) into a student network of similar depth by penalizing a softened version of the teacher’s output. \cite{romero2014fitnets} extended the idea of KD to allow for thinner and deeper student models, using not only the outputs but also the intermediate representations learned by the teacher as hints to improve the training process and final performance of the student.\\
There are several improvements along this direction. \cite{balan2015bayesian} trained a parametric student model to approximate a Monte Carlo teacher. The proposed framework used online training, and used deep neural networks for the student model. Different from previous works which represented the knowledge using the soften label probabilities, \cite{luo2016face} represented the knowledge by using the neurons in the higher hidden layer, which preserved as much information as the label probabilities, but are more compact. \cite{chen2015net2net} accelerated the experimentation process by instantaneously transferring the knowledge from a previous network to each new deeper or wider network. The techniques are based on the concept of function-preserving transformations between neural network specifications. Zagoruyko et al. \cite{zagoruyko2016paying} proposed Attention Transfer (AT) to relax the assumption of FitNet. They transferred the attention maps that are summaries of the full activations.\\
KD-based Approaches can make deeper models thinner and help significantly reduce the computational cost. However, there are a few drawbacks. One is that KD can only be applied to classification tasks with softmax loss function, which limits its usage. Another drawback is that the model assumptions sometimes are too strict to make the performance not so competitive when compared with other types of compression approaches.

\section{Regularization Methods}
Here we give a review of group lasso regularization and its variants for their use of automatically learning a compact structure of a deep network during training in more detail. A general deep network is denoted by
\begin{equation*}
\boldsymbol{y}=f(\boldsymbol{x}; \boldsymbol{\theta})=\sigma\Big(W_l\sigma\big(W_{l-1}\cdots\sigma(W_1\boldsymbol{x}+b_1)\cdots+b_{l-1}\big)+b_l\Big).
\end{equation*}
It can be described as a succession of $L$ layers where the 0th layer is an input layer, the $L$-th layer is an output layer, and others are hidden layers performing linear operations on their input, intertwined with non-linearities defined by activate function $\sigma$. It takes a vector $\boldsymbol{x}$ as input, and returns a scalar $\boldsymbol{y}$ after propagating it through the whole neural network. Each layer $l$ consists of $N_l$ neurons, each of which is encoded by parameters $\theta_l^n=[W_l^n, b_l^n]$, where $W_l^n$ is a linear operator acting on the layer's input and $b_l^n$ is a bias. Altogether, these parameters form the parameter set $\boldsymbol{\theta}=\{\theta_l\}_{1 \leq l \leq L}$, with $\theta_l=\{\theta_l^n\}_{1 \leq n \leq N_l}$.\\
Given a training dataset consisting of N instances ${(\boldsymbol{x}_i, \boldsymbol{y}_i)}_{1 \leq i \leq N}$, the training objective for learning the parameters of the network can be expressed as:
\begin{equation}
    \min_{\boldsymbol{\theta}}\frac{1}{N}\sum_{i=1}^N L(\boldsymbol{y}_i, f(\boldsymbol{x}_i,\boldsymbol{\theta}))+\lambda R(\boldsymbol{\theta})
\label{optimization}
\end{equation}
where $L(\cdot,\cdot)$ is a proper loss function, $R(\cdot)$ is the regularization term and $\lambda$ is the regularization parameter that balances both terms. Standard choices for $L(\cdot,\cdot)$ are the squared error for regression, and the cross-entropy loss for classification.\\
Choosing the correct regularization really impacts the performance of pruning and retraining. The most popular choice of parameter norm penalty is the $l_2$-norm in which the sum of squares of parameters values is added to the loss function: $R_{l_2}(\boldsymbol{\theta})=||\boldsymbol{\theta}||_2$. It is commonly known as 'weight decay' in the network literature. This regularization strategy drives the weights closer to the origin and helps to prevent overfitting since it has an effect of adding a bias term to reduce variance of the model, which in turn results in a lower generalization error. However, the only method to enforce sparsity with weight decay is to artificially force all weights that are lower than a certain threshold in absolute terms to zero. Even in this way, its sparsity effect might be negligible. To obtain a sparse model where large portion of weight is zeroed out, $R(\cdot)$ needs to be a sparsity-inducing regularizer $R_{l_1}(\boldsymbol{\theta})=||\boldsymbol{\theta}||_1$. This $l_1$-norm regularization results in obtaining a sparse weight matrix, since it requires the solution to be found at the corner of the $l_1$-norm ball, thus can eliminate unnecessary elements. The element-wise sparsity can be helpful when most of the features are irrelevant to the learning objective, as in the data-driven approaches. It gives better accuracy after pruning, but before retraining. However, the remaining connections are not as good as with $l_2$ regularization, resulting in slight accuracy reduction after retraining.\\
Since $l_2$-norm has the grouping effect that results in similar weights for correlated features, this will result in complete elimination of some groups, thus removing some input neurons. This has an effect of automatically deciding how many neurons to use at each layer. Still, this group sparsity does not maximally utilize the capacity of the network since there still exists redundancy among the features that are selected. Thus, they chose to apply a sparsity-inducing regularization that obtains a sparse network weight matrix, while also minimizing the redundancy among network weights for better utilization of the network capacity.\\
We propose to do model compression by starting from an over-complete network and canceling the influence of some of its neurons. Note that none of the standard regularizers could achieve this goal: The former favors small parameter values, and the latter tends to cancel out individual parameters, but not complete neurons. In fact, a neuron is encoded by a group of parameters, and the goal therefore translates to making some entire groups go to zero. In \cite{yuan2006model}, the authors introduce group lasso regularization as:
\begin{equation}
    R_{GL}(\boldsymbol{\theta})=\sum_{l=1}^L \sqrt{|p_l|} \sum_{n=1}^{N_l} ||\theta_l^n||_2
\label{GL}
\end{equation}
where $|p_l|$ amounts for the varying group size of the parameters for each neuron in layer $l$ and it ensures that each group gets weighted uniformly. Since the Euclidean norm of a vector $\theta_l$ is zero only if all of its components are zero, this procedure encourages sparsity at both the group and individual levels. That is, $l_2$-norm has the grouping effect that operates on similar weights for correlated features, it will result in  complete elimination of some groups, thus removing some input neurons. This has an effect of automatically deciding how many neurons to use at each layer. That is why group Lasso is an efficient regularization to learn sparse structures, so we could apply it to regularize fully-connected layers (number of neurons and layer depth).\\
The formulation in Eq. \ref{GL} might still be sub-optimal, however, since we lose guarantees of sparsity at the level of single connections among those remaining after removing some of the groups. \cite{friedman2010note} and \cite{simon2013sparse} propose a more general penalty in order to select groups and predictors within a group. They consider the following composite 'sparse group Lasso' (SGL) penalty clearly stated in \cite{scardapane2017group}:
\begin{equation}
    R_{SGL}(\boldsymbol{\theta})=(1-\alpha)R_{GL}(\boldsymbol{\theta})+\alpha R_{l_1}(\boldsymbol{\theta}), 
\label{SGL}
\end{equation}
where $\alpha \in [0,1]$ sets the relative influence of both terms. Note that $\alpha=0$ brings us back to the regularization term of Eq. \ref{GL}.\\
In an elegant paper \cite{alvarez2016learning}, the authors proposed a new algorithm to automatically determine the number of neurons in each layer of the network. They introduced a parameter $\lambda_l$ to measure the influence of the penalty in Eq. \ref{GL}. Specifically, they revised the group lasso regularization term as:
\begin{equation}
    R(\boldsymbol{\theta})=\sum_{l=1}^L \lambda_l \sqrt{|p_l|} \sum_{n=1}^{N_l} ||\theta_l^n||_2.
\label{wGL}
\end{equation}
They said that $\lambda_l$ could be different for various layers and in practice, it was most effective to have two different weights: a relatively small one for the first few layers, and a larger weight for the remaining ones. Similarly, to further leverage this idea within their automatic model selection approach, they extend Eq. \ref{wGL} to a sparse version:
\begin{equation}
    R(\boldsymbol{\theta})=\sum_{l=1}^L \left((1-\alpha) \lambda_l \sqrt{P_l}\sum_{n=1}^N ||\theta_l^n||_2+\alpha  \lambda_l\sum_{n=1}^N |\theta_l^n|\right)
\label{wSGL}
\end{equation}
Learning the right connections is an iterative process. Pruning followed by a retraining is one iteration, and the minimum number of connections could be found after many such iterations. After one such iteration, neurons with zero input connections or zero output connections may be safely pruned and then other connections from that neuron should be furthered removed. Therefore, the retraining phase finally arrives at the result where dead neurons will have both zero input connections and zero output connections. This occurs due to gradient descent and regularization. A neuron that has zero input connections (or zero output connections) will have no contribution to the final results, leading the gradient to be zero for its output connections (or input connections), respectively. When learning terminates, the regularization term will push the parameters of some neurons go to zero in the end. Thus, the dead neurons will be automatically removed during retraining. Furthermore, when considering fully-connected layers, the neurons acting on the output of zeroed-out neurons of the previous layer also become useless so that they can be removed since they have no effect on the output. Ultimately, removing all these neurons yields a more compact architecture than an original, over-complete one.\\

\section{Our Improving Framework of Partial Regularization}
\begin{figure}[H]
\centering
\begin{minipage}[t]{0.48\textwidth}
\centering
\includegraphics[width=6.6cm]{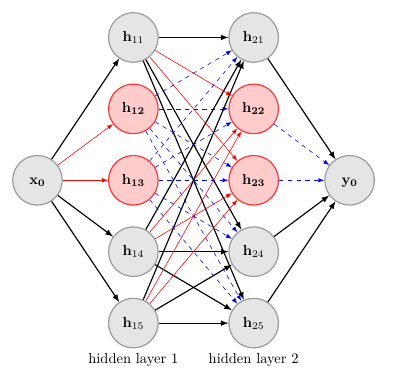}
\end{minipage}
\begin{minipage}[t]{0.48\textwidth}
\centering
\includegraphics[width=6.6cm]{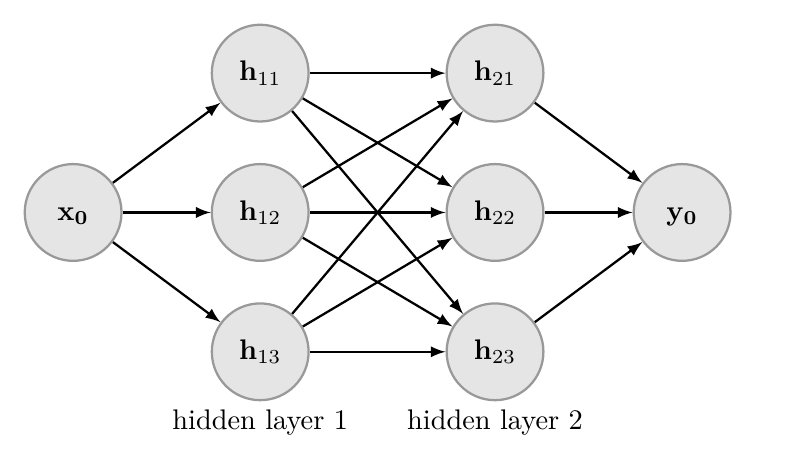}
\end{minipage}
\caption{Original Neural Network(Left), Compact Neural Network(Right).}
\label{procedure}
\end{figure}
Firstly I will describe the process of how node selection works. To make it intuitive,  let's consider the right compact neural network in Fig \ref{procedure} for example. If the norm of outgoing weights from node $h_{11}$ is nearly zero and the norm for that from $h_{12}$ and $h_{13}$ exceeds zero, the output of this layer is essentially based only on $h_{12}$ and $h_{13}$. Under this circumstance, even if we remove node $h_{11}$, it has no large changes and retains the DNN performances that were gained before node pruning. On the other hand, if the norm of ingoing weights to $h_{21}$ is nearly zero and the norm of that to $h_{22}$ and $h_{23}$ for exceeds zero, the output of hidden layer 2 is essentially related to $h_{22}$, $h_{23}$ and the bias terms. In this situation, even if we prune node $h_{21}$, it still has no large changes and the DNN performances accordingly remain almost the same. However, because the node is pruned according to its weight vector norm, its bias remains as a constant. Therefore, we have to remember here to shift the bias for the pruned node to the upper layer nodes. Regardless of the weight vector grouping, if we can control the values of the weight vector norm in the DNN training, or more precisely, selectively minimize the norm values, we can also achieve a mechanism that automatically selects nodes without big changes in the performances of the trained DNNs.\\
In the presence of multiple hidden neurons, permutation invariant property appears, i.e., the model of the observed data is invariant with respect to exchange of arbitrary two hidden neurons. In addition, the permutation invariant property  is a common feature in many modern neural network architectures. It plays an important role in understanding the computation performance of a broad class of neural networks with two or more hidden units.\\
As a matter of fact, we do penalty all if we want to remove all the neurons of one whole layer. Regularization terms in the previous methods contain all the parameters of neural networks. In practice, we usually encounter the application of full regularization since the initial network is artificially set to be really too big. The goal of model compression is to make it as small as possible. In this paper, we don't need to impose penalty on all for the reason that our designed network for each problem is not so large that removing one or more whole hidden layers may have an extremely bad effect to the final results. We propose an improving framework of partial and successive penalty, taking advantage of its permutation invariant property. We give a visual representation of our framework in Fig \ref{procedure}. The original network contains one input layer, two hidden layers and one output layer which takes \textbf{$x_0$} as input and \textbf{$y_0$} as output. Without loss of generality, we can artificially set a series of coefficients in order to remove the particular two neurons of hidden layer 1 that are highlighted in red color. Once the selected neurons are removed, the corresponding outgoing weights represented by blue dashed line should be considered not crucial to the outputs. Therefore, we don't need to penalty all the outgoing connections of hidden layer 1 if we have done the neuron-cut procedure. This is why methods with our improving framework would be faster than previous ones because they require less computation for the regularization terms. Of course, we can also impose the penalty on the input layer since cancelling some input neurons may not cause great changes to the output, unless our network has only one input neuron.\\
Applying the improving framework translates group lasso regularization and its sparse variant into forms of partial regularization. To this end, we introduce a new hyper-parameter $\beta_n$ to modify the group lasso regularization (\ref{GL}):
\begin{equation}
    R(\boldsymbol{\theta})=\sum_{l=1}^L \sqrt{|p_l|} \sum_{n=1}^{N_l} \beta_n ||\theta_l^n||_2
\label{aGL}
\end{equation}
Following the derivation in the last section, we propose to exploit sparse group lasso regularization and rewrite the regularizer as follows:
\begin{equation}
    R(\boldsymbol{\theta})=\sum_{l=1}^L \left((1-\alpha) \sqrt{P_l}\sum_{n=1}^N \beta_n ||\theta_l^n||_2+\alpha \sum_{n=1}^N \beta_n |\theta_l^n|\right)
\label{aSGL}
\end{equation}
\begin{comment}
Moreover, we adopt the idea of adding layer weights in \cite{alvarez2016learning} into the regularization term of Eq. \ref{aSGL} to modify the weighted sparse group lasso method, 
\begin{equation}
    R(\boldsymbol{\theta})=\sum_{l=1}^L \left((1-\alpha) \lambda_l \sqrt{P_l}\sum_{n=1}^N \beta_n ||\theta_l^n||_2+\alpha \lambda_l \sum_{n=1}^N \beta_n |\theta_l^n|\right)
\label{awSGL}
\end{equation}
\end{comment}
However, when we conduct experiments with our improving framework on the group lasso variants adopting layer weights, training performances are better but test performances become worse which means that it doesn't have a better generalization ability. We deduce that it may have a conflict between these two tricks so that we do not combine together.\\
Clearly, to realize our idea, we ccould set the new hyper-parameter $\beta_n$ to be a diagonal matrix with some entries being zeros mathematically. If so, the regularization term contains a smaller number of elements than the corresponding one of full regularization form. What's more, we can set the number of zeros in $\beta_n$ artificially to determine the number of entries left in the regularization terms. Because of the permutation invariant property, it has no effect to the final loss. We will discuss later about the results of different initialization of the hyper-parameter $\beta_n$.\\

\section{Experiments}
In the experiment section, we evaluate our proposal on different datasets with a strong focus on the performance of reducing computation cost and improving network metrics. We begin with a simple toy example to illustrate its general performance and then move on to famous and large real-world datasets which require the use of deeper, larger networks. In particular, we demonstrate the advantages of applying our improving framework with comparison of original methods. Note that hardware and program optimizations can further boost the performance but are not covered in this work.
\subsection{Experiment Setup}
We briefly describe some details of the implementation. The numerical experiments are run on an Acer desktop with 2.60GHz Intel Core i7, and those image classification tasks need one GeForce GTX 1080 Ti GPU for acceleration.\\
\textbf{Datasets:} For testing the performance of regression tasks, we use a toy example of function $y=-x^2$ for 40 points equally distributed in range [-1, 1] and Boston House Pricing Dataset \footnote{UCI Machine Learning Repository}. The latter one was collected in 1978 and each of the 506 entries represent aggregated data about 14 features for homes from various suburbs in Boston, Massachusetts. For testing the performance of classification tasks, we choose Sensorless Drive Diagnosis(SDD) dataset, the extremely famous MNIST dataset and Fashion-MNIST dataset. These data sets include different data types and training set sizes. In the SDD dataset, we wish to predict whether a motor has one or more defective components, starting from a set of 48 features obtained from the motor’s electric drive signals. The dataset is composed of 58508 examples obtained under 11 different operating conditions. The MNIST database is composed of 70 thousands $28\times28$ gray images of the handwritten digits 0-9. It has 60000 images for training and 10000 for test. Fashion-MNIST is a dataset of Zalando's article images-consisting of a training set of 60,000 examples and a test set of 10,000 examples. Each example is a $28\times28$ grayscale image, associated with a label from 10 classes.\\
\textbf{Architecture Design:} We focus on the performance of our partial regularization framework for fully-connected layers. For the architecture of the networks, we artificially set the number of layers to 4, with 1 input layer, 2 hidden layers, 1 output layer and the number of neurons by trial and error. We use Rectified Linear Unit (ReLU) as the activation function for the hidden variables. All the weights in the network are initialized using a normal distribution without any pre-training. We choose $\alpha=0.1$ for the sparse form of methods and fixed regularization factors for different tasks. The size of the mini-batches is set differently depending on the dimensionality of the problem. All of the hyper-parameters are given in Table \ref{table1}, some of which are from \cite{scardapane2017group}.\\ 
\textbf{Training:} We train the neural networks by using the popular Adam \cite{Kingma2014Adam} algorithm with or without mini-batches. Adam is a variant of stochastic gradient descent algorithm, based on adaptive estimates of lower-order moments. In all cases, parameters of the Adam procedure are kept as the default values recommended in the original paper. We also adopted the technique of batch normalization \cite{Ioffe2015Batch}, right after each linear transformation and before activation. This method accelerates the training by allowing a larger step size and easier parameter initialization.\\
\textbf{Implementation Trick:} In the mathematical formula, we introduce a new coefficient $\beta_n$ operating on the regularization terms. According to our proposed method, we choose it to be a diagonal matrix with some entries being 1 and the left ones being 0. If we multiply such a diagonal matrix with the weight matrix, the corresponding several rows of their matrix product will be all zeros and cannot be processed successfully in Tensorflow \cite{tensorflow}. Therefore, we decide to adopt a trick of slicing the parameter matrix only on the first dimension to the required shape. By this way, specific number of parameters are involved to be updated in each iteration, same as the former matrix multiplication idea. For group lasso method and its variants, we assume that each group only has one parameter so that $p_l$ is set to be the width of parameter $\theta_l$. Due to the fluctuation of running time, we choose the average among three tries under the same settings. The following tables contain metrics showing training and test performances, neurons after pruning, sparsity and total running time. Sparsity reports the ratio of the number of parameters that are smaller than a fixed threshold over the total number. \\

\begin{table}[H]
\centering
\caption{Parameters for Neural Networks in the experiments \protect \\
(x/y/z: x-dimensional input layer, y-dimensional hidden layer, and z-dimensional output layer)}
\label{table1}
\begin{tabular}{cccccc}
\toprule
\multirow{2}{*}{Dataset} & \multicolumn{3}{c}{Parameters}\\
\cmidrule(r){2-4}
&Regularization Factor&Neurons&Batch Size \\
\midrule
Boston&0.001&13/40/30/1&-\\
Toy Example&0.001&1/50/50/1&-\\
SDD&0.0001&48/40/40/30/11&500\\
MNIST&0.0001&784/400/300/100/10&400\\
FASHION-MNIST&0.0001&784/400/300/100/10&400\\
\bottomrule
\end{tabular}
\end{table}

\subsection{Regression Tasks}
\begin{figure}[htbp]
\centering
\subfigure[Group Lasso.]{
\includegraphics[width=6.6cm]{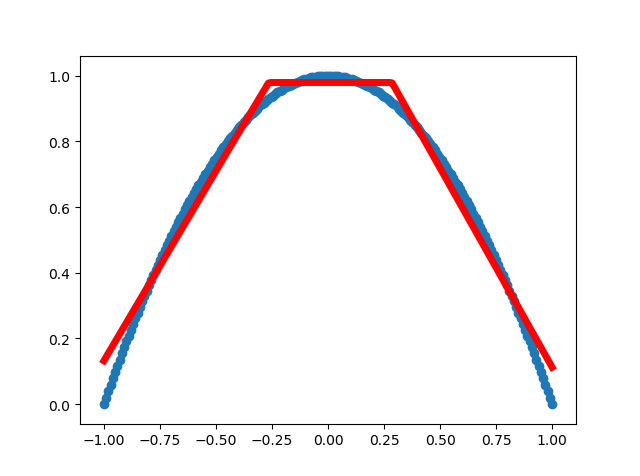}
}
\quad
\subfigure[Improved Group Lasso.]{
\includegraphics[width=6.6cm]{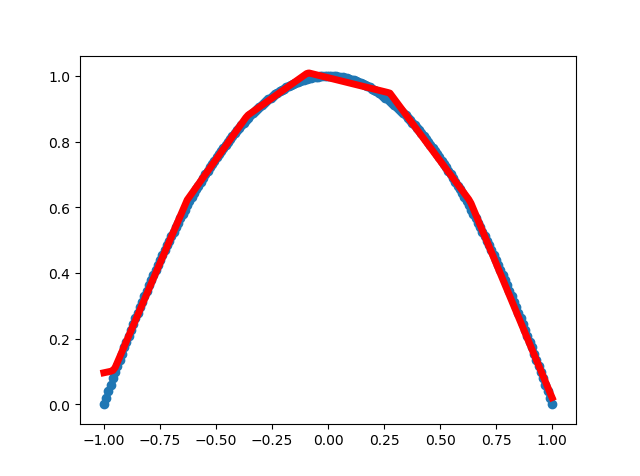}
}
\quad
\subfigure[Sparse Group Lasso.]{
\includegraphics[width=6.6cm]{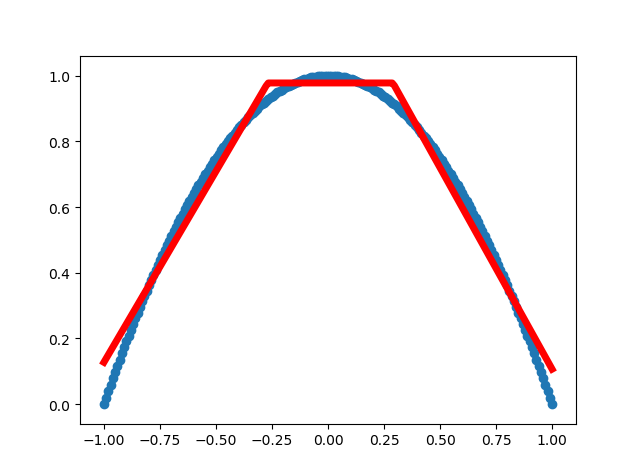}
}
\quad
\subfigure[Improved Sparse Group Lasso.]{
\includegraphics[width=6.6cm]{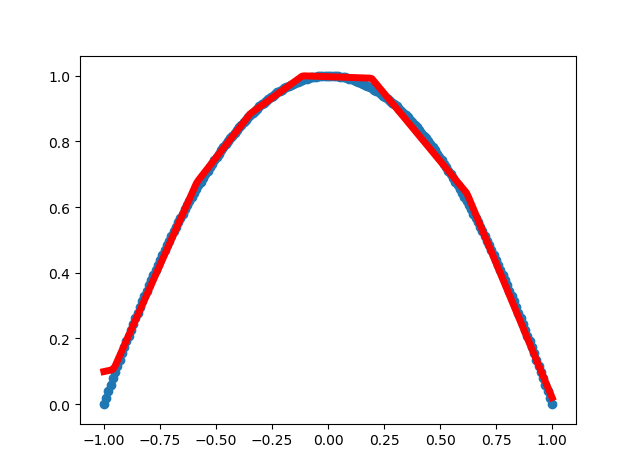}
}
\caption{Blue: Plot of $y=-x^2$ in [-1, 1], Red: Fitting Results}
\label{fit}
\end{figure}

Toy example is a data fitting task using neural networks. Table \ref{table2} shows that methods with our improving framework have lower training and test loss and require less running time. It means that networks after partial regularization can bring better fitting results than those after full regularization more quickly, as shown in Fig. \ref{fit}. Red curves on the right-hand side are closer to the plot of original function.\\

\begin{table}[htbp]
    \centering
    \caption{Results on Toy Example}
    \begin{tabular}{ccccccc}
    \toprule
	Regularization&Training loss&Test Loss&Neurons&Sparsity&Time(s) \\
	\midrule
	GL&1.51E-03&0.36E-02&[1, 30, 2, 1]&[0.96, 0.9988, 0.96]&4.86 \\
	iGL&8.24E-05&0.12E-02&[1, 29, 10, 1]&[0.8, 0.8, 0.8]&4.66 \\
	SGL&1.46E-03&0.33E-02&[1, 49, 2, 1]&[0.96, 0.9988, 0.96]&5.67 \\
	iSGL&1.03E-04&0.14E-02&[1, 50, 10, 1]&[0.8, 0.8, 0.8]&5.13 \\
    \bottomrule
    \end{tabular}
    \label{table2}
\end{table}

\begin{table}[htbp]
    \centering
    \caption{Results on Boston House Pricing Dataset}
    \begin{tabular}{ccccccc}
    \toprule
	Regularization&Training loss&Test Loss&Neurons&Sparsity&Time(s) \\
	\midrule
	GL&1.03E-02&1.05E-02&[12, 3, 1, 1]&[0.9827, 0.9992, 0.9667]&38.10 \\
	iGL&0.99E-02&0.89E-02&[10, 25, 12, 1]&[0.2327, 0.7267, 0.6]&36.77 \\
	SGL&1.02E-02&1.03E-02&[13, 40, 1, 1]&[0.9808, 0.9992, 0.9667]&38.27 \\
	iSGL&1.13E-02&0.89E-02&[13, 35, 10, 1]&[0.2385, 0.7217, 0.6667]&38.13 \\
    \bottomrule
    \end{tabular}
    \label{table3}
\end{table}

Since the main goal of Boston House Pricing task is to construct a working model which has the capability of predicting the value of houses, we naturally separate the dataset into features and the target variable. The features give us quantitative information about each data point and the target variable will be the price we seek to predict. We observe that methods with our improving framework perform better on the training data. Also, partial regularization improves test performances. This means that it provides better predictions of the housing prices with a better model after pruning than methods with full regularization. Additionally, less time is required by applying our improving framework in the experiments. Moreover, it helps to reduce the number of neurons in some layers of the networks.\\
Here we present a result which is not stated but does exist in our experiments. When conducting further training with the obtained slim neural network and comparing the total running steps with those of the original ones, we find that it takes fewer steps to achieve comparable performance. Furthermore, it will be better than that of training original neural network with regularization finally. At the same time, if we train a neural network of the same small size with random initialization, it will always provide us worse performances. That means it's better for us to do model compression first and continue training the slim network with current parameters of the remaining connections.\\

\subsection{Classification Tasks}
We choose a variety of datasets to prove that our partial regularization framework is a general technique for improving the classification performance and generalization ability of neural networks. Since the number of neurons for each layer has great difference, we compute several results for different ratios of zeros in $\beta_n$ for classification tasks and report the values at a particular $\beta_n$ for some metrics to show the effectiveness of our modified methods.\\
\begin{table}[htbp]
    \centering
    \caption{Results on SDD(\textbf{Ratio of Zeros = $\frac{1}{8}$})}
    \begin{tabular}{ccccccc}
    \toprule
    Regularization&Training accuracy&Test accuracy&Neurons&Sparsity&Time(s) \\
	\midrule
	GL&0.9889&0.9874&[47, 25, 11, 10]&[0.59, 0.84, 0.92, 0.71]&507 \\
	iGL&0.9943&0.9928&[48, 20, 8, 15]&[0.55, 0.85, 0.90, 0.65]&489 \\
	SGL&0.9883&0.9875&[47, 22, 9, 9]&[0.67, 0.89, 0.93, 0.74]&528 \\
	iSGL&0.9960&0.9929&[47, 24, 9, 12]&[0.57, 0.79, 0.85, 0.67]&508 \\
    \bottomrule
    \end{tabular}
    \label{table4}
\end{table}

\begin{table}[htbp]
    \centering
    \caption{Results on MNIST(\textbf{Ratio of Zeros = $\frac{1}{8}$})}
    \begin{tabular}{ccccccc}
    \toprule
	Regularization&Training accuracy&Test accuracy&Neurons&Sparsity&Time(s) \\
	\midrule
	GL&0.9598&0.9574&[522, 19, 28, 33]&[0.99, $\approx$ 1.0, 0.99, 0.68]&73 \\
	iGL&0.9722&0.9640&[605, 345, 79, 39]&[0.87, 0.88, 0.88, 0.69]&68 \\
	SGL&0.9569&0.9541&[756, 341, 27, 33]&[0.99, $\approx$ 1.0, 0.99, 0.69]&74 \\
	iSGL&0.9696&0.9592&[759, 384, 81, 46]&[0.87, 0.88, 0.88, 0.70]&74\\
    \bottomrule
    \end{tabular}
    \label{table5}
\end{table}

\begin{table}[htbp]
    \centering
    \caption{Results on FASHION-MNIST(\textbf{Ratio of Zeros = $\frac{1}{4}$})}
    \begin{tabular}{ccccccc}
    \toprule
	Regularization&Training accuracy&Test accuracy&Neurons&Sparsity&Time(s) \\
	\midrule
	GL&0.8528&0.8352&[589, 7, 9, 16]&[$\approx$ 1.0, $\approx$ 1.0, $\approx$ 1.0, 0.84]&328 \\
	iGL&0.9171&0.8577&[748, 260, 84, 27]&[0.76, 0.76, 0.76, 0.74]&307 \\
	SGL&0.8650&0.8467&[771, 304, 10, 17]&[$\approx$ 1.0, $\approx$ 1.0, $\approx$ 1.0, 0.84]&340 \\
	iSGL&0.9131&0.8536&[778, 365, 91, 27]&[0.76, 0.76, 0.76, 0.74]&321 \\
    \bottomrule
    \end{tabular}
    \label{table6}
\end{table}

\begin{figure}[H]
\centering
\begin{minipage}[t]{0.48\textwidth}
\centering
\includegraphics[width=8cm]{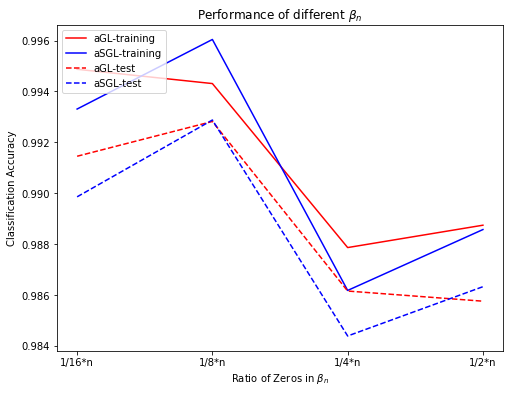}
\end{minipage}
\begin{minipage}[t]{0.48\textwidth}
\centering
\includegraphics[width=8cm]{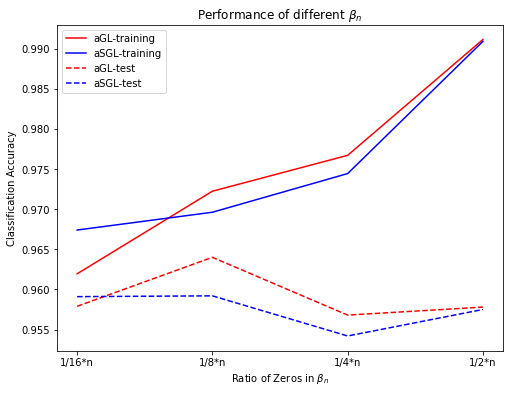}
\end{minipage}
\centering
\begin{minipage}[t]{0.48\textwidth}
\centering
\includegraphics[width=8cm]{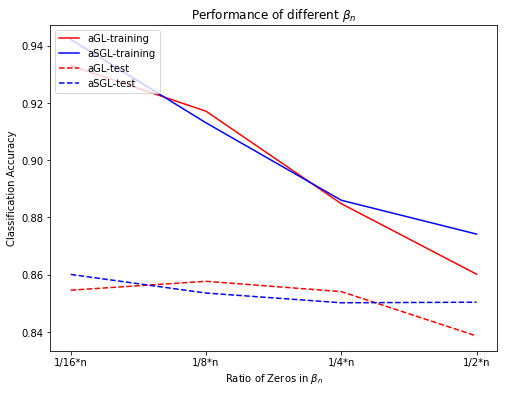}
\end{minipage}
\caption{Classification accuracy of different $\beta_n$ on three datasets: SDD(Upper Left), MNIST(Upper Right), FASHION-MNIST(Lower). }
\label{trend}
\end{figure}
As we can see, no matter which dataset we choose, our partial regularization framework always makes the training better, even with a great improvement under an appropriate choice of the new hyper-parameter $\beta_n$. Similarly, it also has a good impact on test performances. Actually, it is not difficult to find such $\beta_n$ since almost anyone we choose in our experiments could bring better results for most circumstances. Another thing we want to emphasize is that our framework does help to reserve the most important neurons and connections. When compared to the network structure obtained with no regularization, they still get a more compact network with little loss on metrics, sometimes neglected.\\%Additionally, the performance of our framework depends on the value of $\alpha$ in methods of sparse form.\\
When we take a look at the last column of each table, our improving framework brings a minor acceleration on the running time. However, the speed-ups are not such great as we imagine and deduce by mathematical representations. Even though partial regularization contains less parameters which need to be trained, it may owing to that the running time may largely come from data processing steps, background program occupation in Tensorflow when running codes or other unknown reasons. At the same time, since partial regularization reserve more neurons, the computation is more complicated than that of full regularization.\\ 
One main drawback of our framework occurs when considering the performance of parameter reduction. It seems to reserve more parameters than the original ones. We present some thoughts as follows. Maybe the original methods, especially group lasso method, remove so many neurons including some important ones. According to our experiments, we find that neural networks compressed by original full regularization methods contain much less neurons than those of partial regularization methods. If so, much useful information are neglected and cannot be transferred to the following layers. It must result in a bad effect on the performance.\\
What's more, there is no monotone increasing or decreasing trend in test classification accuracy with different $\beta_n$ in Fig. \ref{trend}. Therefore, we could conclude that for any network on a specific dataset, there must exist an optimal zero ratio of $\beta_n$ so that our improving framework could help to get a more compact network along with highly acceptable accuracy and generalization ability. At the same time, artificial settings of arbitrary positions of zeros in $\beta_n$ do not affect the performances of our modified methods so that it verifies the permutation variant property of neural networks conversely. 

\section{Conclusions}
We have introduced an improving framework of partial regularization to accelerate the computations with the help of the permutation invariant property in neural networks and apply it into group lasso method and its sparse variant. To this end, we impose a new coefficient on the parameters of the fully-connected layers in deep neural networks. In mathematical representation, it works within the regularization terms stated in Section 4. Our experiments have demonstrated the advantages of our improving framework on regularization methods for several tasks including those of regression and classification types. We could conclude that if we choose the coefficient $\beta_n$ properly, it has been found useful to improve the training performances and test performances with less or at least equal running time consistently. This means that our partial regularization framework could not only improve the data fitting results, but also increase the generalization ability of compressed neural networks. By the way, it yields a more compact architecture than the initial over-complete network, thus saving both memory and computation costs.\\
The limitation is that we have not shown its benefits on other different architectures, however, we believe it works in any circumstances such as convolutional and recurrent neural networks. We will go further study on this aspect. In addition, to deal with the aforementioned drawback of more reserved parameters, we could think of applying another modification on partial regularization framework in a proper way to obtain an optimal network structure with acceptable performance, such as the elegent one in \cite{yoon2017combined}. %Further, element-wise sparsity, while achieving a memory-efficient model, usually do not result in meaningful speedups in practical network architectures such as CNNs, since the bottleneck is in the convolutional operations that do not reduce much when the number of filters stays the same. 

\section*{Acknowledgement}
The authors thank Yingzhou Li and Li Zhou for helpful discussions and insightful comments.

\bibliographystyle{unsrt}
\nocite{*}
\bibliography{references}  %%% Remove comment to use the external .bib file (using bibtex).

\begin{thebibliography}{10}

\bibitem{ochiai2017automatic}
Tsubasa Ochiai, Shigeki Matsuda, Hideyuki Watanabe, and Shigeru Katagiri.
\newblock Automatic node selection for deep neural networks using group lasso
  regularization.
\newblock In {\em 2017 IEEE International Conference on Acoustics, Speech and
  Signal Processing (ICASSP)}, pages 5485--5489. IEEE, 2017.

\bibitem{cheng2017survey}
Yu~Cheng, Duo Wang, Pan Zhou, and Tao Zhang.
\newblock A survey of model compression and acceleration for deep neural
  networks.
\newblock {\em arXiv preprint arXiv:1710.09282}, 2017.

\bibitem{lecun1990optimal}
Yann LeCun, John~S Denker, and Sara~A Solla.
\newblock Optimal brain damage.
\newblock In {\em Advances in neural information processing systems}, pages
  598--605, 1990.

\bibitem{hassibi1993second}
Babak Hassibi and David~G Stork.
\newblock Second order derivatives for network pruning: Optimal brain surgeon.
\newblock In {\em Advances in neural information processing systems}, pages
  164--171, 1993.

\bibitem{srinivas2015data}
Suraj Srinivas and R~Venkatesh Babu.
\newblock Data-free parameter pruning for deep neural networks.
\newblock {\em arXiv preprint arXiv:1507.06149}, 2015.

\bibitem{han2015learning}
Song Han, Jeff Pool, John Tran, and William Dally.
\newblock Learning both weights and connections for efficient neural network.
\newblock In {\em Advances in neural information processing systems}, pages
  1135--1143, 2015.

\bibitem{chen2015compressing}
Wenlin Chen, James Wilson, Stephen Tyree, Kilian Weinberger, and Yixin Chen.
\newblock Compressing neural networks with the hashing trick.
\newblock In {\em International Conference on Machine Learning}, pages
  2285--2294, 2015.

\bibitem{HanMao16}
Song Han, Huizi Mao, and William~J. Dally.
\newblock Deep compression: Compressing deep neural networks with pruning,
  trained quantization and huffman coding.
\newblock {\em arXiv preprint arXiv:1510.00149}, 2015.

\bibitem{ullrich2017soft}
Karen Ullrich, Edward Meeds, and Max Welling.
\newblock Soft weight-sharing for neural network compression.
\newblock {\em arXiv preprint arXiv:1702.04008}, 2017.

\bibitem{lebedev2016fast}
Vadim Lebedev and Victor Lempitsky.
\newblock Fast convnets using group-wise brain damage.
\newblock In {\em Proceedings of the IEEE Conference on Computer Vision and
  Pattern Recognition}, pages 2554--2564, 2016.

\bibitem{zhou2016less}
Hao Zhou, Jose~M Alvarez, and Fatih Porikli.
\newblock Less is more: Towards compact cnns.
\newblock In {\em European Conference on Computer Vision}, pages 662--677.
  Springer, 2016.

\bibitem{wen2016learning}
Wei Wen, Chunpeng Wu, Yandan Wang, Yiran Chen, and Hai Li.
\newblock Learning structured sparsity in deep neural networks.
\newblock In {\em Advances in neural information processing systems}, pages
  2074--2082, 2016.

\bibitem{li2016pruning}
Hao Li, Asim Kadav, Igor Durdanovic, Hanan Samet, and Hans~Peter Graf.
\newblock Pruning filters for efficient convnets.
\newblock {\em arXiv preprint arXiv:1608.08710}, 2016.

\bibitem{lebedev2014speeding}
Vadim Lebedev, Yaroslav Ganin, Maksim Rakhuba, Ivan Oseledets, and Victor
  Lempitsky.
\newblock Speeding-up convolutional neural networks using fine-tuned
  cp-decomposition.
\newblock {\em arXiv preprint arXiv:1412.6553}, 2014.

\bibitem{tai2015convolutional}
Cheng Tai, Tong Xiao, Yi~Zhang, Xiaogang Wang, et~al.
\newblock Convolutional neural networks with low-rank regularization.
\newblock {\em arXiv preprint arXiv:1511.06067}, 2015.

\bibitem{denil2013predicting}
Misha Denil, Babak Shakibi, Laurent Dinh, Nando De~Freitas, et~al.
\newblock Predicting parameters in deep learning.
\newblock In {\em Advances in neural information processing systems}, pages
  2148--2156, 2013.

\bibitem{sainath2013low}
Tara~N Sainath, Brian Kingsbury, Vikas Sindhwani, Ebru Arisoy, and Bhuvana
  Ramabhadran.
\newblock Low-rank matrix factorization for deep neural network training with
  high-dimensional output targets.
\newblock In {\em 2013 IEEE international conference on acoustics, speech and
  signal processing}, pages 6655--6659. IEEE, 2013.

\bibitem{srivastava2014dropout}
Nitish Srivastava, Geoffrey Hinton, Alex Krizhevsky, Ilya Sutskever, and Ruslan
  Salakhutdinov.
\newblock Dropout: a simple way to prevent neural networks from overfitting.
\newblock {\em The Journal of Machine Learning Research}, 15(1):1929--1958,
  2014.

\bibitem{szegedy2017inception}
Christian Szegedy, Sergey Ioffe, Vincent Vanhoucke, and Alexander~A Alemi.
\newblock Inception-v4, inception-resnet and the impact of residual connections
  on learning.
\newblock In {\em Thirty-First AAAI Conference on Artificial Intelligence},
  2017.

\bibitem{wu2017squeezedet}
Bichen Wu, Forrest Iandola, Peter~H Jin, and Kurt Keutzer.
\newblock Squeezedet: Unified, small, low power fully convolutional neural
  networks for real-time object detection for autonomous driving.
\newblock In {\em Proceedings of the IEEE Conference on Computer Vision and
  Pattern Recognition Workshops}, pages 129--137, 2017.

\bibitem{hinton2015distilling}
Geoffrey Hinton, Oriol Vinyals, and Jeff Dean.
\newblock Distilling the knowledge in a neural network.
\newblock {\em arXiv preprint arXiv:1503.02531}, 2015.

\bibitem{romero2014fitnets}
Adriana Romero, Nicolas Ballas, Samira~Ebrahimi Kahou, Antoine Chassang, Carlo
  Gatta, and Yoshua Bengio.
\newblock Fitnets: Hints for thin deep nets.
\newblock {\em arXiv preprint arXiv:1412.6550}, 2014.

\bibitem{balan2015bayesian}
Anoop~Korattikara Balan, Vivek Rathod, Kevin~P Murphy, and Max Welling.
\newblock Bayesian dark knowledge.
\newblock In {\em Advances in Neural Information Processing Systems}, pages
  3438--3446, 2015.

\bibitem{luo2016face}
Ping Luo, Zhenyao Zhu, Ziwei Liu, Xiaogang Wang, and Xiaoou Tang.
\newblock Face model compression by distilling knowledge from neurons.
\newblock In {\em Thirtieth AAAI Conference on Artificial Intelligence}, 2016.

\bibitem{chen2015net2net}
Tianqi Chen, Ian Goodfellow, and Jonathon Shlens.
\newblock Net2net: Accelerating learning via knowledge transfer.
\newblock {\em arXiv preprint arXiv:1511.05641}, 2015.

\bibitem{zagoruyko2016paying}
Sergey Zagoruyko and Nikos Komodakis.
\newblock Paying more attention to attention: Improving the performance of
  convolutional neural networks via attention transfer.
\newblock {\em arXiv preprint arXiv:1612.03928}, 2016.

\bibitem{yuan2006model}
Ming Yuan and Yi~Lin.
\newblock Model selection and estimation in regression with grouped variables.
\newblock {\em Journal of the Royal Statistical Society: Series B (Statistical
  Methodology)}, 68(1):49--67, 2006.

\bibitem{friedman2010note}
Jerome Friedman, Trevor Hastie, and Robert Tibshirani.
\newblock A note on the group lasso and a sparse group lasso.
\newblock {\em arXiv preprint arXiv:1001.0736}, 2010.

\bibitem{simon2013sparse}
Noah Simon, Jerome Friedman, Trevor Hastie, and Robert Tibshirani.
\newblock A sparse-group lasso.
\newblock {\em Journal of Computational and Graphical Statistics},
  22(2):231--245, 2013.

\bibitem{scardapane2017group}
Simone Scardapane, Danilo Comminiello, Amir Hussain, and Aurelio Uncini.
\newblock Group sparse regularization for deep neural networks.
\newblock {\em Neurocomputing}, 241:81--89, 2017.

\bibitem{alvarez2016learning}
Jose~M Alvarez and Mathieu Salzmann.
\newblock Learning the number of neurons in deep networks.
\newblock In {\em Advances in Neural Information Processing Systems}, pages
  2270--2278, 2016.

\bibitem{Kingma2014Adam}
Diederik Kingma and Jimmy Ba.
\newblock Adam: A method for stochastic optimization.
\newblock {\em Computer Science}, 2014.

\bibitem{Ioffe2015Batch}
Sergey Ioffe and Christian Szegedy.
\newblock Batch normalization: accelerating deep network training by reducing
  internal covariate shift.
\newblock In {\em International Conference on International Conference on
  Machine Learning}, 2015.

\bibitem{tensorflow}
Martín Abadi, Ashish Agarwal, Paul Barham, Eugene Brevdo, Zhifeng Chen, Craig
  Citro, G.s Corrado, Andy Davis, Jeffrey Dean, Matthieu Devin, Sanjay
  Ghemawat, Ian Goodfellow, Andrew Harp, Geoffrey Irving, Michael Isard,
  Yangqing Jia, Rafal Jozefowicz, Lukasz Kaiser, Manjunath Kudlur, and
  Xiaoqiang Zheng.
\newblock Tensorflow: Large-scale machine learning on heterogeneous distributed
  systems.
\newblock 03 2016.

\bibitem{yoon2017combined}
Jaehong Yoon and Sung~Ju Hwang.
\newblock Combined group and exclusive sparsity for deep neural networks.
\newblock In {\em Proceedings of the 34th International Conference on Machine
  Learning-Volume 70}, pages 3958--3966. JMLR. org, 2017.

\end{thebibliography}
%%% and comment out the ``thebibliography'' section.

\end{document}